\documentclass[runningheads,a4paper]{llncs}

\usepackage{amssymb}
\usepackage{graphicx}

\usepackage{url}
\urldef{\mailsa}\path|luo@ms.k.u-tokyo.ac.jp|    
\newcommand{\keywords}[1]{\par\addvspace\baselineskip
\noindent\keywordname\enspace\ignorespaces#1}

\begin{document}

\mainmatter  

\title{Reinterpreting the Transformation Posterior in Probabilistic Image Registration}

\titlerunning{\scriptsize{Reinterpreting the Transformation Posterior in Probabilistic Image Registration}}

%
%
\author{Jie Luo$^1$, Karteek Popuri$^2$,  Dana Cobzas$^3$, Hongyi Ding$^4$ \\ and Masashi Sugiyama$^{1,4}$}
\authorrunning{J.Luo, K.Popuri, D.Cobzas, H.Ding, and M.Sugiyama}

\institute{$^1$Graduate School of Frontier Sciences, The University of Tokyo, Japan\\ $^2$School of Engineering Science, Simon Fraser University, Canada\\ $^3$Computing Science Department, University of Alberta, Canada \\ $^4$Department of Computer Science, The University of Tokyo, Japan
\mailsa 
}

%
%

\toctitle{Lecture Notes in Computer Science}
\tocauthor{Authors' Instructions}
\maketitle

\vspace{-5mm}
\begin{abstract}

\emph{Probabilistic image registration methods estimate the posterior distribution of transformation. The conventional way of interpreting the transformation posterior is to use the mode as the most likely transformation and assign its corresponding intensity to the registered voxel. Meanwhile, summary statistics of the posterior are employed to evaluate the registration uncertainty, that is the trustworthiness of the registered image. Despite the wide acceptance, this convention has never been justified. In this paper, based on illustrative examples, we question the correctness and usefulness of conventional methods. In order to faithfully translate the transformation posterior, we propose to encode the variability of values into a novel data type called ensemble fields. Ensemble fields can serve as a complement to the registered image and a foundation for developing advanced methods to characterize the uncertainty in registration-based tasks. We demonstrate the potential of ensemble fields by pilot examples.}
\end{abstract}
\vspace{-7mm}

\section{Introduction}

Since many medical tasks are based on non-rigid image registration, the trustworthiness of registered images, which is also known as the registration uncertainty, is considered critical. Due to factors such as the high degree of freedom in the non-rigid transformation model, the presence of homogeneous intensity regions and the variability of human anatomy, it is insufficient to only report a unique transformation for non-rigid registration. In order to characterize the registration uncertainty, recent non-rigid registration methods have been adapted to a probabilistic framework that estimates the posterior distribution of transformation \cite{Cobzas}\cite{Simpson}\cite{Risholm}\cite{Popuri}\cite{Andrews}\cite{Heinrich}.

Probabilistic non-rigid registration methods can be broadly categorized into discrete probabilistic registration (DPR) and continuous probabilistic registration (CPR). The transformation posteriors estimated by DPR and CPR have different forms. DPR discretizes the transformation space into a set of displacement vectors. Then it uses discrete optimization techniques, such as a graph-based approach, to compute a categorical distribution as the transformation posterior \cite{Cobzas}\cite{Popuri}\cite{Andrews}\cite{Heinrich}. CPR is essentially a Bayesian registration framework, with the estimated transformation posterior given by a multivariate continuous distribution \cite{Simpson}\cite{Risholm}.
\vspace{-3mm}

\subsubsection{Related Work}The conventional way of interpreting the transformation posterior is to use the mode as the most likely transformation and assign its corresponding intensity to the registered voxel. Subsequently, summary statistics of the posterior are employed to evaluate the registration uncertainty. Various summary statistics have been used in the probabilistic registration literature. The Shannon entropy was used to measure the registration uncertainty of DPR \cite{Lotfi}. Meanwhile, the variance \cite{Simpson}, standard deviation \cite{Simpson2}, inter-quartile range \cite{Risholm} and covariance Frobenius norm \cite{Wassermann} of the transformation posterior were used to quantify the registration uncertainty of CPR. In order to visually assess the registration uncertainty, each of these summary statistics was either mapped to a color scheme, or an object overlaid on the registered image. By inspecting the color or the object's geometry, clinicians can infer the trustworthiness of the appearance for the registered image.

In the past few years, researchers almost exclusively used the above convention to interpret the transformation posterior. Despite the wide acceptance, this convention has never been justified. Is it appropriate to assign the corresponding intensity of the posterior mode to the registered voxel? Do those summary statistics truly give insight on the trustworthiness of the registered image? In the following sections, based on illustrative examples, we question the correctness and usefulness of conventional methods. In order to faithfully translate the transformation posterior, we propose to encode the variability of values by a novel data type called \emph{ensemble fields}. Ensemble fields can serve as a valuable complement to the registered image and a foundation for developing advanced methods to characterize the uncertainty in registration-based tasks. We also demonstrate its potential using pilot examples.
\vspace{-3mm}

\section{Potentially Critical Issues in Conventional Methods}
\vspace{-1mm}

Conventional methods interpret the transformation posterior by:
(1) Using the posterior mode as the most likely transformation and assigning its corresponding intensity to the registered voxel. 
(2) Employing summary statistics of the transformation posterior to evaluate the registration uncertainty.
Based on illustrative examples, we question the correctness of (1) and the usefulness of (2).

 For the convenience of illustration, we use RWIR as the probabilistic registration scheme in all examples \cite{Cobzas}\cite{Popuri}\cite{Andrews}. In the RWIR setting, let $I_\mathrm{f}$ and $I_\mathrm{m}$ respectively be the fixed and moving image $I_\mathrm{f}  ,I_\mathrm{m}: \Omega_I\rightarrow\mathbb{R},\Omega_I\subset\mathbb{R}^d, d=2\: or \: 3$. RWIR discretizes the transformation space into a set of $K$ displacement vectors $\mathcal{D} = \{\mathbf{d}_k\}_{k=1}^K, \mathbf{d}_k\in\mathbb{R}^d$. These displacement vectors radiate from voxels on $I_\mathrm{f}$ and point to their candidate corresponding locations on $I_\mathrm{m}$. For every voxel $v_i$, the algorithm computes a unity-sum probabilistic vector $\mathcal{P}(v_i)=\{P_k(v_i)\}_{k=1}^K$ as the transformation posterior. $P_k(v_i)$ is the corresponding probability of displacement vector $\mathbf{d}_k$. The most likely transformation $\mathbf{d}_\mathrm{m}$ for $v_i$ is the displacement in $\mathcal{D}$ that has the highest probability in $\mathcal{P}(v_i)$. 
\vspace{-3mm}

\subsection{Correctness}

Probabilistic registration methods estimate a transformation posterior. Conventionally, researchers impose the intensity corresponding to the most likely transformation, which is the posterior mode, on the registered voxel as the most likely intensity. However, does the corresponding intensity of the most likely transformation always equal to the most likely intensity estimated by the posterior?
 \vspace{-8mm}

\begin{figure}
	\centering
	\includegraphics[height=2.9cm]{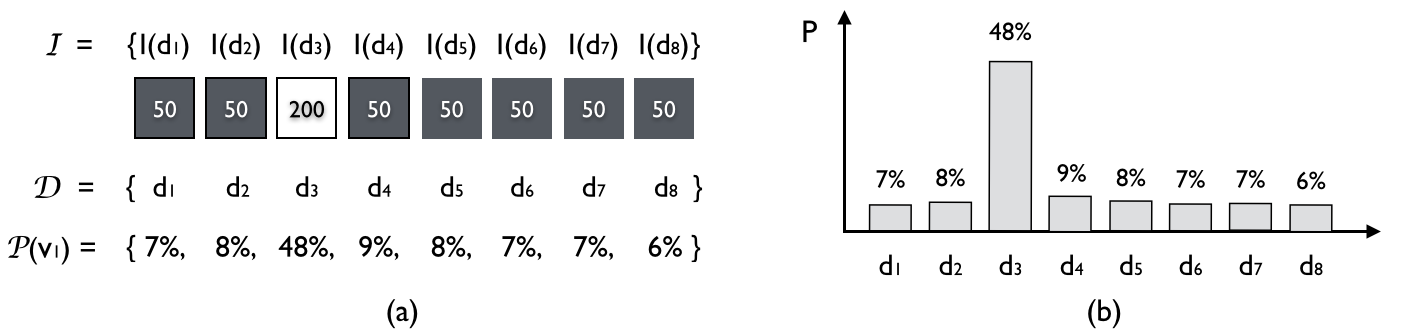}
	 \vspace{-7mm}
	\caption{(a)The RWIR posterior in the hypothetical setting; (b)A bar chart illustrating the transformation posterior.}
	\label{fig:correct1}
\end{figure}
 \vspace*{-6mm}
 
In a hypothetical setting, assuming $v_1$ on $I_f$ is the voxel we want to register. As shown in Fig.1(a), the transformation $\mathcal{D} = \{\mathbf{d}_k\}_{k=1}^8$ is a set of 8 displacement vectors. $\mathcal{P}(v_1)=\{P_k(v_1)\}_{k=1}^8$ is the posterior of $\mathcal{D}$. The corresponding intensity values, which are intensities of those corresponding locations on $I_\mathrm{m}$, of all displacement vectors in $\mathcal{D}$ are stored in $\mathcal{I}=\{I(\mathbf{d_k})\}_{k=1}^8$. For clarity, suppose that there are only two different intensity values in $\mathcal{I}$, one is 50, and the other is 200. The color of squares in Fig.1(a) indicates the appearance of that intensity value. Fig.1(b) is a bar chart illustrating the transformation posterior. We can observe that $\mathbf{d}_3$ is the most likely transformation $\mathbf{d}_m$. Conventionally, the corresponding intensity of the most likely transformation $I(\mathbf{d}_\mathrm{m})=I(\mathbf{d_3})=200$ will be assigned to the registered $v_1$. 

\begin{figure}[t]
	\centering
	\includegraphics[height=3.3cm]{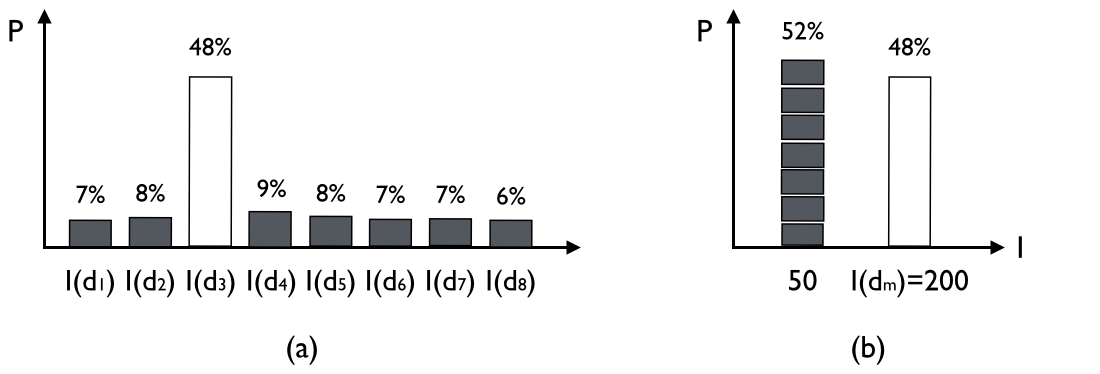}
	\vspace{-4mm}
	\caption{(a)A bar chart of the transformation posterior taking into account $I(\mathbf{d_k})$. The color of each bar indicates the appearance of $I(\mathbf{d_k})$; (b)Intensity histogram of the registered $v_1$.}
	\vspace{-5mm}
	\label{fig:correct2}
\end{figure}

The probability of $\mathbf{d}_3$ is considerably higher than that of other displacement vectors. Based on the common sense, the intensity of registered $v_1$ should be trustworthy. However, if we take into account the intensity value $I(\mathbf{d_k})$ associated with each $\mathbf{d_k}$, and form an intensity distribution from the transformation posterior, as shown in Fig.2, it is clear that $I(\mathbf{d_m})=200$ is not the most likely intensity. Displacement vectors in $\mathcal{D} = \{\mathbf{d}_k\}_{k\in\{1,\ldots,8\}\backslash\{3\}}$ are not the most likely transformation, yet their combined corresponding intensities outweigh $I(\mathbf{d}_3)$.
 
This counter intuitive result implies that the corresponding intensity of the most likely transformation can differ from the most likely intensity estimated by the posterior. Extreme cases like the above example may not happen in practice, yet we can still question this convention whether it is reasonable to disregard the intensity value associated with the transformation posterior. 

More precisely, in a probabilistic registration setting, the transformation $R_T$ and the intensity value $R_I$ are both regarded as random variables. Imposing the corresponding intensity of the most likely transformation on the registered voxel is equivalent to regarding the mode of $R_T$ as if it is the mode of $R_I$. Even $R_T$ and $R_I$ are intuitively correlated to each other, this slightly reckless approach is questionable.

\vspace{-3mm}
\subsection{Usefulness}

After assigning the intensity of the most likely transformation to every registered voxel, it is again a convention to employ summary statistics of the transformation posterior to evaluate the registration uncertainty, that is the trustworthiness of the registered image. In most cases, these summary statistics are mapped to a color scheme and overlaid on the registered image. By interpreting colors of registered voxels, clinicians can infer the trustworthiness of the registration result. Such a convention is useful in a sense that it can quickly draw clinicians' attention to regions with high transformation uncertainty. However, does high transformation uncertainty always indicate high registration uncertainty?

In another hypothetical setting, assuming $v_2$ on $I_f$ is the voxel we want to register. As shown in Fig.3(a), the transformation $\mathcal{D} = \{\mathbf{d}_k\}_{k=1}^4$ is a set of 4 displacement vectors. $\mathcal{P}(v_2)=\{P_k(v_2)\}_{k=1}^4$ is the posterior of $\mathcal{D}$. The corresponding intensity values of all displacement vectors in $\mathcal{D}$ are stored in $\mathcal{I}=\{I(\mathbf{d_k})\}_{k=1}^4$. Since $\mathbf{d}_3$ is the most likely transformation, $I(\mathbf{d_3})$ will be assigned to the registered $v_2$. For a categorical posterior, the conventional way of evaluating the transformation uncertainty is using the Shannon entropy. In this example, the Shannon entropy $E(\mathcal{P}(v_2))\approx2$. Given the size of $\mathcal{P}(v_2)$, $2$ is a considerably large entropy, hence the registration uncertainty of $v_2$ is suggested high by the conventional method. However, once again, we take into account the intensity value $I(\mathbf{d_k})$ associated with each displacement vector $\mathbf{d_k}$. As shown on Fig.3(b), even $\mathbf{d_1},\mathbf{d_2},\mathbf{d_3}$ and $\mathbf{d_4}$ are different displacement vectors, they correspond to the same intensity value. By the generated intensity histogram in Fig.3(c), we can see  
the only possible intensity value for $v_2$ is in fact equal to $I(\mathbf{d_3})$. Therefore, the appearance of the registered voxel is quite trustworthy which contradicts with the outcome of the conventional method. 

\begin{figure}
	\centering
	\includegraphics[height=2.8cm]{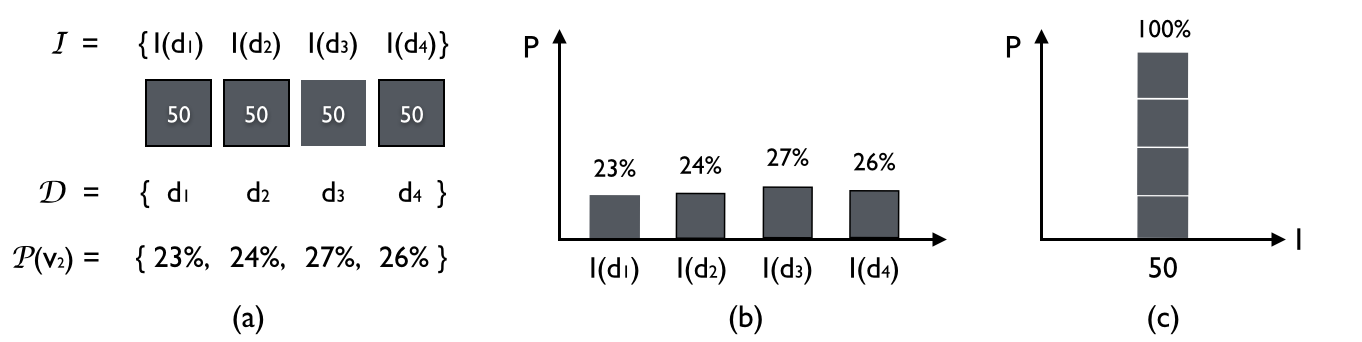}
	\vspace{-3mm}
	\caption{(a)The RWIR posterior for the hypothetical example; (b)A bar chart of the transformation posterior taking into account $I(\mathbf{d_k})$. The color of each bar indicates the appearance of $I(\mathbf{d_k})$; (c)Intensity histogram of the registered $v_2$.}
		\vspace{-5mm}
	\label{fig:useful}
\end{figure}

The above example implies that high transformation uncertainty does not guarantee high registration uncertainty. In practice, homogeneous intensity regions, such as the center of a tumor, sometimes can "fool" the registration algorithm to estimate diverse transformation. Due to using transformation uncertainty to evaluate the registration uncertainty, conventional methods are prone to false results for those regions \cite{Simpson}\cite{Lotfi}. In a probabilistic registration setting, since the transformation $R_T$ and the intensity value $R_I$ are considered random variables, employing Shannon entropy $E(\mathcal{P}(v))$ of the transformation posterior to indicate the trustworthiness of the intensity is similar to using the $E(\mathcal{P}(v))$ of $R_T$ to infer the variance of $R_\mathrm{I}$, which is obviously ineffective.

 \vspace{-3mm}
\section{Reinterpreting the Transformation Posterior}

The conventional interpretation of the transformation posterior has two drawbacks:
(1) Using summary statistics to give a quick and simple description of the transformation posterior and therefore disregarding the rich information that may impact the subsequent registration-base tasks;
(2) Inferring the registration uncertainty solely by the transformation uncertainty and overlooking the influence of other variables, such as the intensity.

In order to faithfully translate the transformation posterior and eventually characterize the uncertainty of registration-based tasks, we reinterpret the transformation posterior by encoding the variability of values into a novel data type called \emph{ensemble fields}.

 \vspace{-3mm}
\subsection{Introducing Ensemble Fields}

 Ensemble fields are a special type of volume data that is used to store possible outcomes of a numerical simulation \cite{Obermaier}.
 \vspace{-5mm}

 \begin{figure}
 	\centering
 	\includegraphics[height=2.6cm]{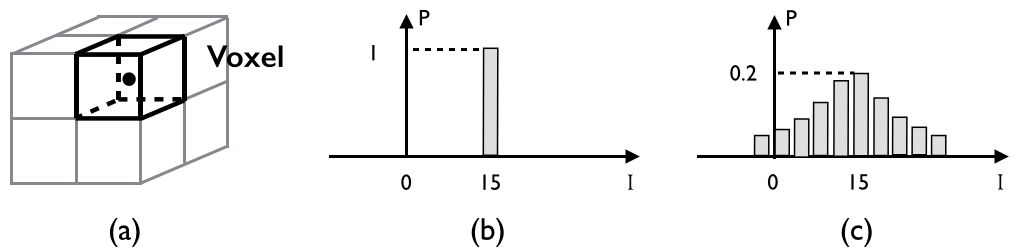}
 	 \vspace{-3mm}
 	\caption{(a) An out-lined voxel; (b) Only a single scalar value is stored in voxels of regular volume data; (c) An intensity distribution is stored in voxels of scalar ensembles.}
 	 \vspace{-5mm}
 	\label{fig:compare}
 \end{figure}

  Let $\mathcal{A}$ be the type of values that can be $\mathbb{R}^n, \mathbb{R}^{m\times n}$ as well as categorical. In an ensemble field, each voxel corresponds to a random variable $R\in\mathcal{A}$. To account for variability, $N$ realizations of $R$, forming a distribution, are stored in every voxel of an ensemble field. Based on the type of stored values, ensemble fields can be categorized into scalar ensemble fields, vector ensemble fields, etc. If a volume data stores intensity distribution, like those in Fig.2(b) and Fig.3(c), then it is a scalar ensemble field. Fig.4 illustrates the difference between a voxel of regular volume data and that of a scalar ensemble field. At the current stage, we construct ensemble fields by aggregating the probability over each realization. In the future, we can also add regularization to meet higher level demands.

 \vspace{-3mm}
\subsection{The Potential of Ensemble Fields}
 
The goal of this paper is to introduce ensemble fields so that researchers can use it as a foundation and develop advanced method to characterize the uncertainty of registration-based tasks. In this section, we give four pilot examples explaining scenarios when utilizing ensemble fields is useful.
  \vspace{-2mm}
 
 \begin{figure}[t]
 	\centering
 	\includegraphics[height=5.0cm]{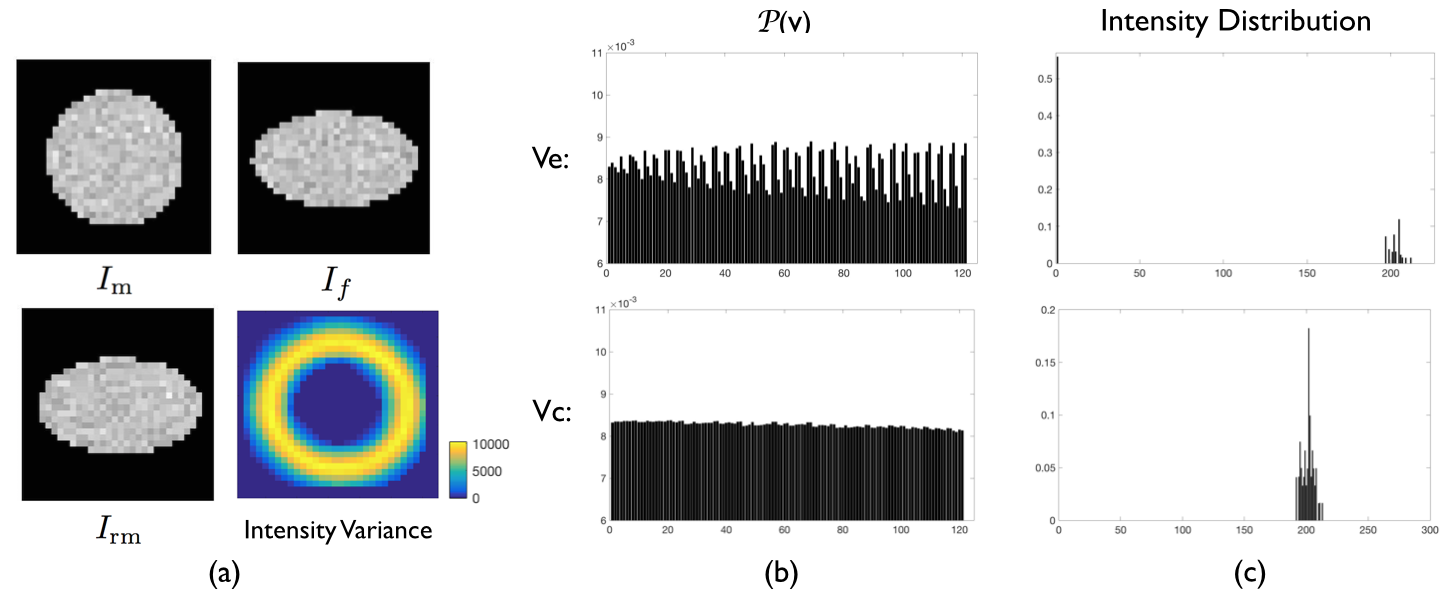}
 	\vspace{-4mm}
 	\caption{(a) Input and result of the circle-ellipse example; (b) Bar charts illustrating the transformation posterior of $v_e$ and $v_c$; (c) Intensity distributions of $v_e$ and $v_c$ being stored in the scalar ensemble.}
 	\vspace{-5mm}
 	\label{fig:compare}
 \end{figure}
 
\paragraph{Estimating the trustworthiness of registered images} Estimating the registration uncertainty is the most straightforward application of ensemble fields. This task can be done by measuring the variance of the generated scalar ensemble fields.
In an example with synthetic 2D images, we registered a circle to an ellipse by a RWIR having 121 displacement vectors. To give more insight, we take a close look at two voxels, $v_e$ near the edge of the eclipse on the registered moving image $I_\mathrm{rm}$, and $v_c$ at the center of the eclipse. As can be seen from Fig.5(b), the transformation posteriors of $v_c$ is more uniformly distributed than that of $v_e$. Conventionally, $v_c$ is likely to be reported having higher registration uncertainty than $v_e$. However, once we construct a scalar ensemble field and measure the intensity variance, which can be inferred from Fig.5(c), it turns out that the registered $v_e$ is less trustworthy than $v_c$. In fact, as shown in Fig.5(a), voxels with high registraton unceratinty are all located near the edge of the ellipse.
	 \vspace{-2mm}

  \paragraph{Computer aided diagnosis} Accurate segmentation labels are important for computer aided diagnosis (CAD). One viable automatic segmentation strategy is the probabilistic atlas-image registration, by which the labels on the atlas are propagated to voxels on the image. Using label ensemble fields, we can account the uncertainty of label propagation, and thus efficiently capture the subject variability. In the example shown in Fig.6(a), we propagate the ventricle label from $I_\mathrm{m}$ to $I_\mathrm{f}$. By RWIR, we can obtain a label distribution for every voxel on $I_\mathrm{f}$ and compute a probability map of the ventricle label. Results are presented in Fig.6(b) and Fig.6(c) respectively. Brighter colors indicate higher probabilities.
  
\vspace{-1mm}

\begin{figure}[t]
	\hspace{-2mm}
	\begin{minipage}[b]{0.51\linewidth}
		\includegraphics[width=1\linewidth]{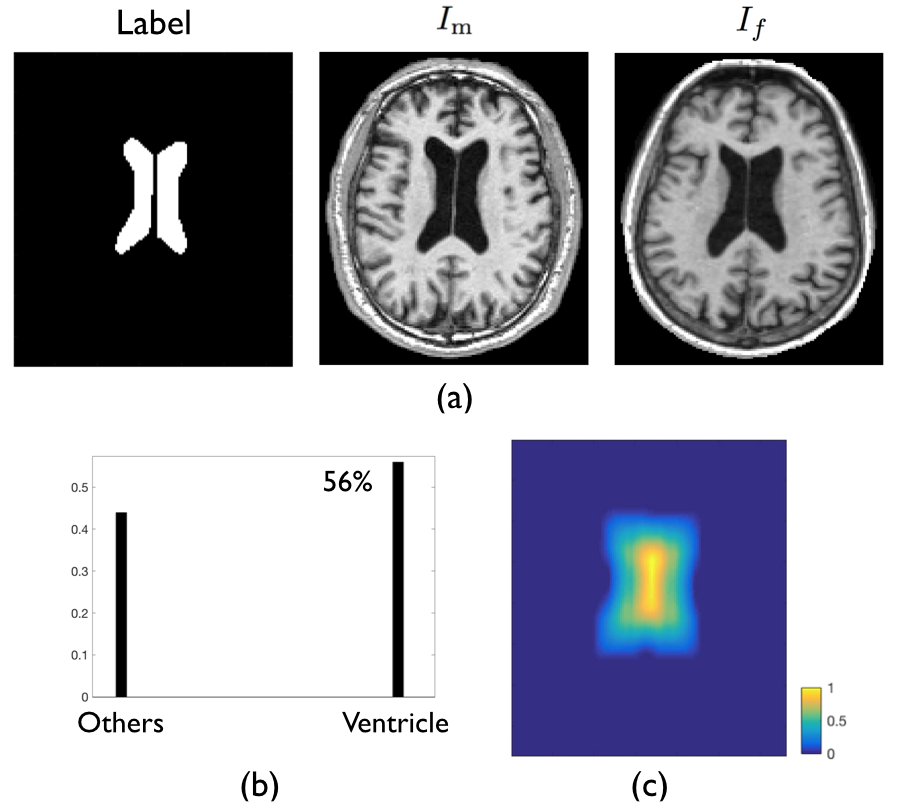}
		\label{fig:prob1_6_2}	
	\end{minipage}
	\begin{minipage}[b]{0.48\linewidth}
		\includegraphics[width=1\linewidth]{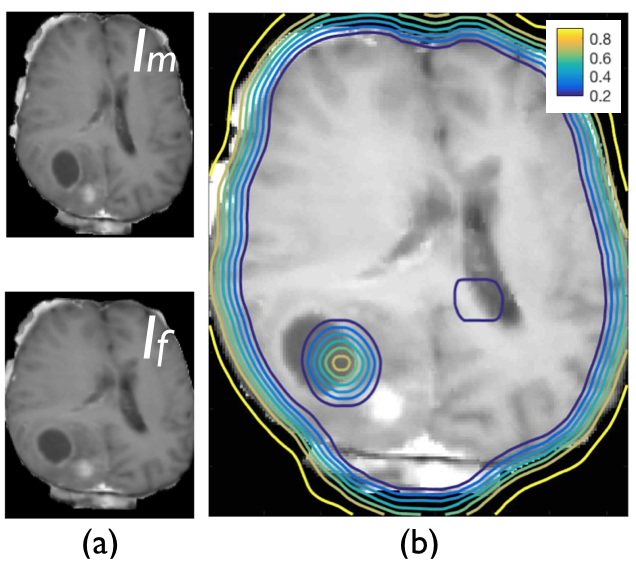}
		\label{fig:prob1_6_1}
	\end{minipage}
	
	\vspace*{-8mm}
	
	\begin{minipage}[t]{.49\linewidth}
		\centering
		\caption{(a)The ventricle label,  $I_\mathrm{m}$ and $I_\mathrm{f}$; (b) The label distribution of a voxel; (c) The probability color map of the generated label ensemble.}
	\end{minipage}%
	\hspace{2mm}
	\begin{minipage}[t]{.48\linewidth}
		\centering
		\caption{(a) The original image $I_\mathrm{m}$ and the distorted image $I_\mathrm{f}$; (b)Intensity iso-contours as the possible tumor boundaries.}
	\end{minipage}%
	\vspace*{-5mm}
\end{figure}

\paragraph{Visualization} Visualization is another promising field to utilize ensemble fields. For instance, when clinicians investigate the boundary of a tumor on the registered image, they often focus on voxels near the tumor boundary. If the conventional color coding of these voxels indicates low uncertainty, then the boundary they are currently seeing is trustworthy. In case these voxels' uncertainty is suspected high, it should be made clear that how the tumor boundary may vary according to transformation posterior. Conventional summary statistics based visualization methods discard the intensity information and can not depict possible appearance changes of the image. One way to achieve this is using scalar ensemble fields. In Fig.7(a), we distorted the image $I_\mathrm{m}$ and set it to $I_\mathrm{f}$. After generating the scalar ensemble field, we thresholded on a manually set intensity value and displayed its iso-contours as the possible boundaries of the registered tumor. In practice, this method is not quite feasible. Alternatively, researchers in \cite{Risholm}\cite{Risholm2} would outline the tumor on $I_\mathrm{m}$ and build a histogram volume by sampling the posterior. Their method is essentially transforming $I_\mathrm{m}$ into a binary labeled image, and thresholding on a disguised label ensemble field. In addition, if we are interested in visualizing the uncertainty of derived features, such as the possible trajectories of registered fiber tracks, generating a vector ensemble field may be the right start.
	 \vspace{-1mm}

\paragraph{Validation} Some research have reported it was beneficial to utilize information in the transformation posterior. However, additional findings from our experiments revealed that the information in the posterior is not always useful. In the circle ellipse example shown in Fig.5, we found some voxels whose most likely intensity does not equal to the corresponding intensity of the posterior mode $I(\mathbf{d}_\mathrm{m})$. Unexpectedly, $I(\mathbf{d}_\mathrm{m})$ is the one that is closer to the ground truth. How did the posterior give worse result? By ensemble fields, the transformation posterior is \textquotedblleft exposed\textquoteright\textquoteright, we can investigate the correlation between variables, such as the transformation and intensity, and develop some standards to evaluate the credibility of the transformation posterior.

 \vspace{-3mm}
\section{Discussion}

With the advance of technologies and the increasing complexity of medical tasks, utilizing the full transformation posterior will become more prevalent. In order to develop a principled way to interpret the result of probabilistic registration, we feel it is necessary to share our findings and introduce assets, such as ensemble fields, to the research community. Based on pilot examples, we demonstrate the usefulness of ensemble fields in various registration-based tasks. In the future, we will investigate the credibility of the transformation posterior and develop learning schemes that take ensemble fields as input for CAD.


\vspace{-3mm}
\begin{thebibliography}{4}



\bibitem{Cobzas} Cobzas, D., Sen, A.: Random Walks for Deformable Registration. In: MICCAI. LNCS, vol. 6892, pp. 557--565. Springer, Toronto (2011)
\bibitem{Simpson} Simpson, I.J.A., Schnabel, J.A., Norton, I., Groves, A.R., Andersson, J.L.R., Woolrich, M.W. : Probabilistic Inference of Regularisation in Non-rigid Registration. NeuroImage. 59, 2438-2451 (2012)


\bibitem{Risholm} Risholm, P., Janoos, F., Norton, I., Golby, A.J., Wells III, W.M.: Bayesian Characterization of Uncertainty in Intra-subject Non-rigid Registration. Med. Image Anal. 17(5), 538-555 (2013)


\bibitem{Popuri} Popuri, K., Cobzas, D., Jagersand, M.: A Variational Formulation for Discrete Registration. In: MICCAI. LNCS, vol. 8151, pp. 187--194. Springer, Nagoya (2013)


\bibitem{Andrews} Andrews, S., Tang, L., Hamarneh, G.: Topology Preservation and Anatomical Feasibility in Random Walker Image Registration. In: MICCAI. LNCS, vol. 8673,
pp. 210--217. Springer, Boston (2014)

\bibitem{Heinrich} Heinrich, M.P., Simpson, I.J.A., Papiez, B.W., Brady, M.: Deformable Image Registration by Combining Uncertainty Estimates From Supervoxel Belief Propagation. Med. Image Anal. 27, 57-71 (2016)

\bibitem{Simpson2} Simpson, I.J.A., Cardoso, M.J., Norton, I., Modat, M., Woolrich, M.W., Andersson, J.L.R, Schnabel, J.A., Ourselin, S.: Probabilistic Non-linear Registration with Spatially Adaptive Regularisation. Med. Image Anal. 26, 203-216 (2015)


\bibitem{Lotfi} Lotfi, P., Tang, L., Andrews, S., Hamarneh, G.: Improving Probabilistic Image Registration via Reinforcement Learning and Uncertainty Evaluation. In: MLMI. LNCS, vol. 8184, pp. 187--194. Springer, Nagoya (2013)

\bibitem{Wassermann} Wasserman, D., Toews, M., Niethammer, M, Wells III, W.M.: Probabilistic Diffeomorphic Registration: Representing Uncertainty. In: WBIR. LNCS, vol. 8545, pp. 72--82. Springer, London (2014)



\bibitem{Risholm2} Risholm, P., Balter, J., Wells III, W.M.: Estimation of Delivered Dose in Radiotherapy: The influence of Registration Uncertainty. In: MICCAI. LNCS, vol. 6891, pp. 548--555. Springer, Toronto (2011)


\bibitem{Obermaier} Obermaier, H., Kenneth, I.J.: Future Challenges for Ensemble Visualization. IEEE. TCGA. 34(3), 8--11 (2014)









\end{thebibliography}
\end{document}